\documentclass[letterpaper, 10 pt, conference]{ieeeconf}  

\IEEEoverridecommandlockouts                              

\usepackage{amsmath} 
\usepackage{graphicx}
\usepackage{booktabs}
\usepackage{amsmath}
\usepackage{url}
\usepackage{hyperref}

\newcommand{\norm}[2]{\left \lVert #1 \right \rVert^2_2}
\newcommand{\argmin}[1]{\underset{#1}{\operatorname{arg}\,\operatorname{min}}\;}

\newcommand{\norms}[1]{\left \lVert #1 \right \rVert}

\title{\LARGE \bf
MLP-SLAM: Multilayer Perceptron-Based Simultaneous Localization and Mapping
}

\author{Taozhe Li$^{1}$ and Wei Sun$^{2}$
\thanks{$^{1}$Taozhe Li is a graduate stduent at the School of Aerospace and Mechanical Engineering,
        University of Oklahoma, Norman, OK, US.
        {\tt\small Taozhe.Li-1@ou.edu}}%
\thanks{$^{2}$Wei Sun is an Assistant Professor at the School of Aerospace and Mechanical Engineering, University of Oklahoma, Norman, OK, US.
        {\tt\small wsun@ou.edu}}%
}

\begin{document}

\maketitle
\thispagestyle{empty}
\pagestyle{empty}

\begin{abstract}
The Visual Simultaneous Localization and Mapping (V-SLAM) system has advanced significantly in recent years, achieving high precision in environments with few dynamic objects. However, its performance declines in settings with numerous movable objects, such as those with pedestrians, cars, and buses, which are common in outdoor scenes. While some researchers are tackling this issue with artificial mathematical methods, these often suffering less generality and information loss. To address this problem, we propose a novel Multilayer Perceptron (MLP)-based method that utilizes complete geometric information to prevent information loss and capable to apply in various situations. To best of our knowledge, this is the first time that machine learning-based method introduced for distinguishing dynamic and static feature points in the SLAM field. Additionally, there is currently no publicly available dataset for evaluating dynamic and static feature classification methods. To fill this gap, we have created a publicly available dataset containing over 50,000 feature points. Experimental results show that our MLP-based dynamic and static feature point discriminator outperforms other methods on this dataset. Furthermore, our proposed MLP-based real-time stereo SLAM system achieves the highest average precision and fastest speed on the outdoor KITTI Odometry and KITTI360 datasets compared to other dynamic SLAM systems. The open-source code and datasets will be accessible at https://github.com/TaozheLi/MLP-SLAM upon paper acceptance.

\end{abstract}


\section{INTRODUCTION}
\begin{figure*}
      \centering
      \includegraphics[width=1.0\textwidth, height=0.25\textheight]{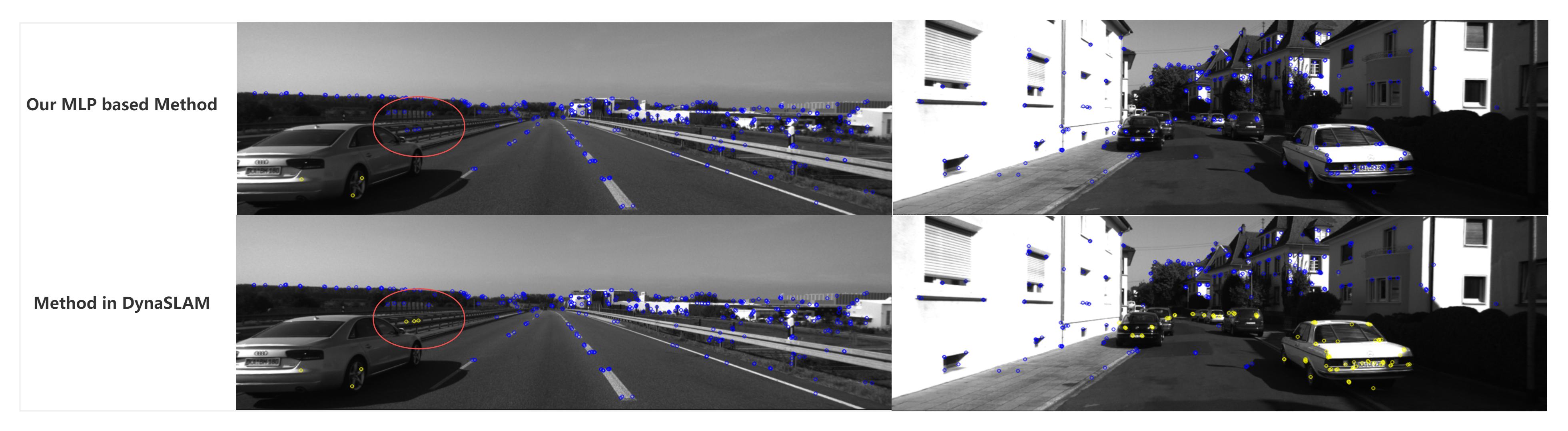}
      \caption{Qualitative comparison between our method and method in DynaSLAM. The static feature point is represented by a blue circle, while the dynamic feature point is represented by a yellow circle. The top image displays the results of our method, while the bottom image shows the results of the method in DynaSLAM. In the left image, the method in DynaSLAM categorizes feature points on the highway guardrail as dynamic, whereas our method correctly predicts them as static. In the right image, numerous static feature points associated with parked cars on the roadside are misclassified as dynamic by the method in DynaSLAM, whereas our method correctly identifies them as static.}
      \label{figurelabel1}
   \end{figure*}

Visual Simultaneous Localization and Mapping (V-SLAM) allows for precise ego-localization and accurate mapping using a lightweight, cost-effective camera sensor. It is vital in robotics, space exploration, and autonomous driving, where it enhances navigation and decision-making. However, most existing SLAM systems are heavily reliant on the assumption of a static world, including state-of-the-art (SOTA) systems such as ORB-SLAM2 \cite{ORB-SLAM2}, ORB-SLAM3 \cite{ORB-SLAM3}, LSD-SLAM \cite{LSD-SLAM} and Direct Sparse Odometry \cite{engel2017direct}. These systems typically perform with high precision in environments with predominantly static objects. Nevertheless, their precision tends to deteriorate when deployed in dynamic environments, which are common in real world scenarios. Some algorithms perform well in dynamic scenes with the utilization of robust kernels like Huber kernel \cite{chebrolu2021adaptive}, but it relies heavily on the accuracy of the depth measurement obtained from LiDAR sensor. Research efforts aimed at addressing the degradation of SLAM systems in dynamic environments can be categorized into three types. Type I focuses only on geometric methods, which involve classifying feature points based solely on geometric information. This information includes epipolar error, reprojection error, image intensity, depth, and optical flow, as demonstrated in works such as \cite{zhu2022nice}. Type II, exemplified by works like DynaSLAM \cite{DynaSLAM}, leverages prior class information obtained from object detection models or semantic segmentation models to identify features within bounding boxes or masks of objects belonging to dynamic classes as outliers, which are then discarded during camera pose estimation. While these approaches partially mitigate the degradation of SLAM systems in dynamic environments, their effectiveness is limited. For instance, features extracted from stationary parked cars on the roadside may be erroneously removed, despite their static nature and potential value in estimating camera pose, leading to information loss. In Figure \ref{figurelabel1}, a qualitative comparison is presented between our Multilayer Perceptron (MLP)-based discriminator for dynamic and static feature points and the method proposed in DynaSLAM, using a snapshot from the KITTI Odometry dataset \cite{geiger2013visionKITTIDatasets}. The top image displays the results of our method, while the bottom image shows the results of the method in DynaSLAM. In the left image, the method in DynaSLAM categorizes feature points on the highway guardrail as dynamic, whereas our method correctly predicts them as static. In the right image, numerous static feature points associated with parked cars on the roadside are misclassified as dynamic by the method in DynaSLAM, whereas our method correctly identifies them as static. This evidence supports the effectiveness and reliability of our method. Type III represents the current mainstream approach, which combines methods from Type I and Type II, as demonstrated in works such as DynaSLAM II \cite{DynaSLAMII}, CFP-SLAM \cite{CFP-SLAM}, PointSLOT \cite{PointSLOT}, VDO-SLAM \cite{VDO-SLAM}, and D-SLAM \cite{wen2023dynamic}, by initially acquiring prior class information through deep learning models such as object detection models or semantic segmentation models. The YOLO series is the most common deep learning model in the SLAM domain, known for its high precision and speed, such as YOLOv5 \cite{jocher2022ultralytics} and YOLOACT \cite{bolya2019yolact}. Subsequently, geometric methods are used to eliminate dynamic feature points by leveraging prior model information. These methods consider factors such as epipolar error, re-projection error, image intensity, depth, and velocity. However, many methodologies in Type III rely on only partial geometric information. For instance, some methods solely focus on epipolar error \cite{yu2018ds}, re-projection error \cite{PointSLOT}, or a combination thereof \cite{CFP-SLAM}. These approaches are commonly employed in the dynamic SLAM field. Nevertheless, these methods often require manual parameter tuning, lack generality, and are excessively laborious to fine-tune. Furthermore, the evaluation of methods distinguishing between dynamic and static feature points is typically conducted indirectly through SLAM field metrics, as there is a lack of publicly available datasets specifically designed for this classification task. This indirect evaluation method tends to be inaccurate and unreliable due to the randomness in the Random Sample Consensus (RANSAC) process in camera pose estimation within the SLAM system.

To address the aforementioned gaps, we introduce a MLP-based real-time stereo SLAM system and establishes publicly available datasets for evaluating dynamic and static feature points method directly. The contributions outlined in this work are as follows:
\begin{itemize}
    \item We introduce an open-source MLP-based real-time stereo system designed to enhance performance in dynamic environments. This system leverages comprehensive geometry information and eliminates the need for manual parameter configuration. To the best of our knowledge, this is the first time a machine learning-based method has been introduced to distinguish dynamic and static feature points in the SLAM field. 
    \item We have constructed publicly available datasets comprising more than 50,000 feature points sourced from the KITTI Odometry datasets \cite{geiger2013visionKITTIDatasets} for the purpose of training and testing dynamic and static feature points discriminators. The efficacy of it can be evaluated directly with this dataset.
    \item We elaborate our methods 's performance in all.
\end{itemize}

\section{Related Work}
\subsection{Dynamic SLAM}
DynaSLAM has gained significant attention in Visual Simultaneous Localization and Mapping since its release. It identifies feature points within the semantic masks of potential moving objects as outliers and discards them. Although it improves performance in the KITTI Odometry dataset, it may misrepresent dynamic and static objects, as shown in Figure \ref{figurelabel1}. Furthermore, DynaSLAM experiences high latency due to its dependence on the computationally intensive MaskRCNN model \cite{he2017mask} for semantic segmentation. Subsequent research endeavors have aimed to address the degradation of SLAM systems in dynamic environments. For instance, work CFP-SLAM\cite{PointSLOT} introduced a Moving Objects Recognition (MOR) approach to differentiate between dynamic and static feature points by utilizing the re-projection error. In comparison, work \cite{CFP-SLAM} employs a DBSCAN\cite{ester1996densityDBSCAN} clustering machine learning technique to filter out features within bounding boxes that belong to the background and then employ a geometric method that incorporates geometric information such as re-projection error and epipolar error to distinguish dynamic and static feature points. An inherent limitation in \cite{PointSLOT} and \cite{CFP-SLAM} is the requirement of the establishment and tuning of manual parameters, which restricts the application scenarios and diminishes the generality of the SLAM system. Furthermore, these approaches do not fully leverage geometric information. The VDO-SLAM\cite{VDO-SLAM} and DynaSLAM II\cite{DynaSLAMII} are are extensions of DynaSLAM, with DynaSLAM II representing the current state-of-the-art (SOTA) in camera motion estimation. Both works initially estimate the camera pose using feature points outside the potential dynamic class semantic mask, akin to the approach in DynaSLAM. Subsequently, they estimate the object pose for each frame and optimize the camera and object poses simultaneously through Bundle Adjustment (BA) factor graph. However, these methodologies encounter high latency issues due to data association challenges between dynamic objects and object pose estimation. Additionally, the precision of object pose estimation is compromised by the limited number of matched feature points and the quality of feature matching for each object.

\subsection{Multilayer Perceptron}
Multilayer Perceptron (MLP) \cite{rosenblatt1961principles} is commonly utilized as a tool for classification tasks, and it has been effectively employed in artificial intelligence, particularly in the field of computer vision. Some renowned applications include the utilization of MLP to classify 3d point cloud models \cite{qi2017pointnet} and for natural language processing \cite{vaswani2017attention}. In the field of SLAM, studies \cite{sucar2021imap, zhu2022nice} apply the MLP to decode grid features into occupancy values in the mapping part, study \cite{mildenhall2021nerf} uses the MLP to implicitly represent feature points in space. To the best of our knowledge, the method introduced in our letter is the first instance of applying the MLP as a discriminator for dynamic and static feature points in the SLAM domain.

\section{Overview}
\subsection{Notations in SLAM}
World and camera coordinates are denoted by \{$W$\} and \{$C$\} respectively. The $i_\text{th}$ feature point is denoted by $x_i$, and its depth is represented by $d_i$. The coordinate of feature point $x_i$ in camera and world coordinate are describe by $^{t}X^{B}_{i}$, $B \in \{W, C\}$. Moreover, $^{t}P_{i}$ marks the location of feature point $x_i$ in the image plane, and the operation $\pi(.)$ represents projecting an 3D point cloud from camera coordinates into the image plane. i.e. $^{t}P_{i}=\pi(^{t}X^{C}_{i})$. Besides, the translation matrix $^{t}T_{CW} = \begin{bmatrix}^{t}R_{CW}, \textbf{t}_{CW}\\0,1 \end{bmatrix} \in SE(3)$ is used to describe a transformation from the world coordinate $\{W\}$ to the camera coordinate $\{C\}$. i.e. $^{t}X^{C}_{i} = {^{t}R_{CW}}{^{t}X^{W}_{i}} +{t}_{CW} $ . The essential matrix between $t-1$ frame and $t$ frame is denoted by $^{t}E$. 

\subsection{Camera Pose Estimation}
In the camera pose estimation process, we search the static feature point correspondences among consecutive frames to avoid the impact of dynamic objects. 
The camera pose is acquired through minimizing the re-projection error of static feature point correspondences, which is described by:
\begin{equation}
^{t}T_{CW} = \argmin{ {^{t}T^{*}_{CW}} }\sum_{i}^{N}{{\rho(\norm{^{t}e_{i}}))}}, \label{eq:minerror}
\end{equation}
where
\begin{equation}
^{t}e_{i} = {^{t}}P_{i} - \pi(^{t}T_{CW} {{^{t-1}}\hat{X}^{W}_{i}}),
\end{equation}
and the operation $\rho (\cdot)$ is a robust kernel function for reducing impact of feature points with big error, and $N$ denotes the number of static feature correspondences. The $^{t-1}\hat{X}^{W}_{i}$ indicates homogeneous coordinate of ${^{t-1}X^{W}_{i}}$.

\subsection{Notations in MLP}
The generated input feature for each feature point $x_i$ is represented by $v_i \in R^3$. Each generated feature $v_i$ consists of three parts, which are $e_I$, $e_D$ and $e_{Re}$. i.e. $v_i = \left \{\begin{matrix} e_I \\ e_D \\ e_{Re} \end{matrix} \right\} $. The $e_I$, $e_D$ and $e_{Re}$ represent the image intensity error, epipolar error and re-projection error respectively. They are computed as follows
\begin{align}
e_I({^{t-1}x_{i}}, ^{t}x_{i}) &= \norm{I({^{t-1}P_{i}})-I({^{t}P_{i}})}{2}, \label{eq:error1} \\
e_D(^{t}E, {^{t-1}x_{i}}, ^{t}x_{i})&=\frac{{{^{t}P_{i}}}^{T} {^{t}E}}{{\norms{{{^{t}P_{i}}} {^{t}E}}_{2}}} \cdot  {^{t-1}P_{i}}, \label{eq:error2} \\
e_{Re}(R_{}, t_{}, {^{t-1}x_{i}}, ^{t}x_{i})&=\left \lVert \pi(R_{} \pi({^{t}P_{i}}, d_i)^{-1} + t_{}) - {^{t-1}P_{i}}  \right \rVert^2_2.  \label{eq:error3}
\end{align}
We use operation $M(\cdot)$ to denote applying MLP model for classifying dynamic and static feature points. The prediction result  denoted by $y_i \in R^2$. And the class of each feature point is denoted by $c_i$. For our problem, the class $c_i$ equals $0$ or $1$. We used $0$ to indicate static class and $1$ for dynamic class. Thus, the MLP model predicted process can be described by $c_i = argmax(y_{i})$,
where $y_i = M(v_i)$.

\subsection{System Structure}
\begin{figure*}
      \centering
      \includegraphics[width=0.9\textwidth, height=0.2\textheight]{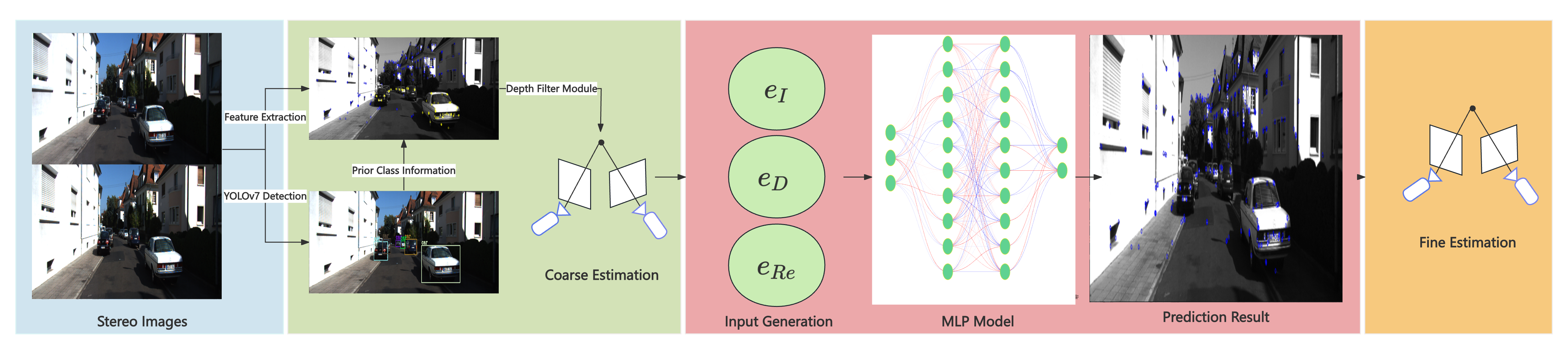}
      \caption{Diagram of our proposed MLP based real-time stereo SLAM system. It was built on ORB-SLAM2. There are four main parts in total. The blue section involves pre-processing, feature extraction, object detection, and object tracking. The pre-processing and feature extraction parts are default setting of ORB-SLAM2.The green section is for coarse estimation, generating a rough pose estimation. The pink section involves an MLP model for classifying dynamic and static feature points within potential dynamic objects. The yellow section is for fine estimation, refining the camera pose generated in the coarse estimation stage by minimizing the re-projection error of all static feature points. These points include those outside the bounding box of potential dynamic objects, filtered by the Depth Filter Module, and classified as static by the MLP model.}
      \label{figurelabel2}
   \end{figure*}

\begin{figure*}
      \centering
      \includegraphics[width=0.9\textwidth, height=0.15\textheight]{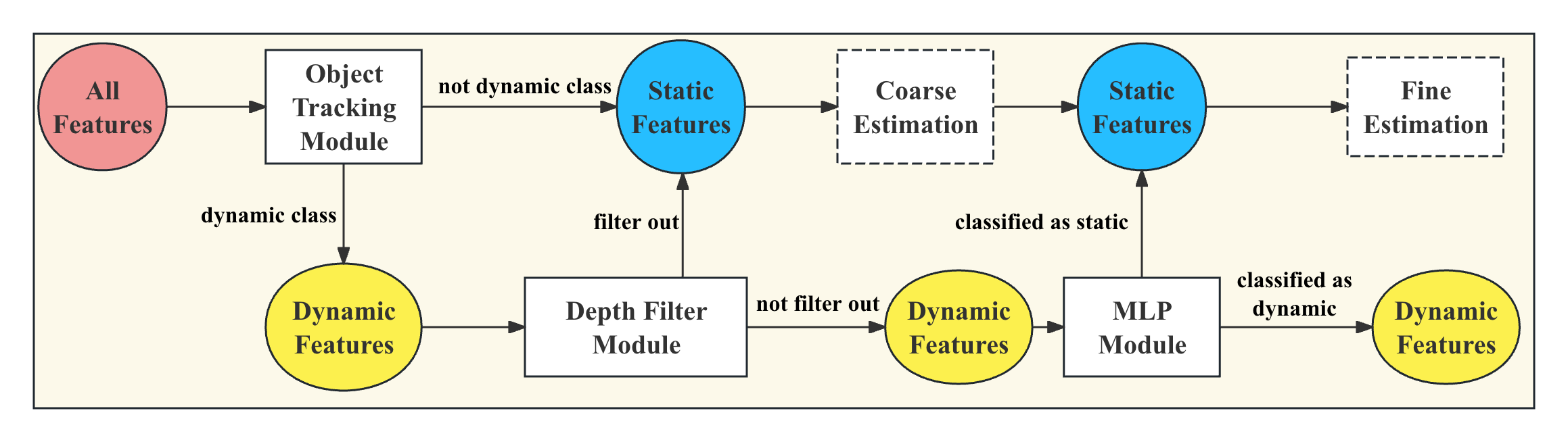}
      \caption{Feature points flow among different module of our SLAM system. The yellow circle represented dynamic feature points set, and the blue one standed for static feature points set. Moreover, all square represented an module of our SLAM system, the one outline with dashed line indicated that it has no impact to feature point sets.}
      \label{figurelabel-d}
   \end{figure*}

The diagram of our MLP-based stereo real-time SLAM system is shown in Figure \ref{figurelabel2}. It is built on ORB-SLAM2 with an extra dynamic and static object discriminator module to filter dynamic feature points. Initially, the input stereo images undergo ORB feature extraction. Subsequently, object detection and tracking are performed on the left image. The process then moves to a coarse estimation stage where features outside the bounding box of potential dynamic objects, and the background static features within the bounding box that are filtered by the Depth Filter Module (DFM), are used for coarse camera pose estimation. The DFM will be explained in detail in Section \ref{subsection:depth}. Then each feature point $x_i$ that is inside the bounding box and not filtered by the DFM generates an input feature $v_i$ using Eq. (\ref{eq:error1}), (\ref{eq:error2}), and (\ref{eq:error3}). An example of the bounding box is shown in Figure \ref{figurelabel2} as the green box around the car. The MLP model assigns a label $c_i$ to each feature point to distinguish static and dynamic feature points. Finally, all static feature points are used for fine camera pose estimation. Feature point flows across the different modules in our method are shown in Figure \ref{figurelabel-d} .


\section{Proposed Method}

In this section, details of each module and our collected dynamic and static feature points classification dataset will be elaborated. 

\subsection{Depth Filter Module (DFM)} \label{subsection:depth}
We opt for object detection and tracking models over semantic segmentation models in our SLAM system to reduce latency, as the latter typically requires more computational resources. However, object detection models produce less accurate bounding boxes compared to mask from semantic segmentation models. To address this issue, we introduce the DFM to remove background features within the bounding box, classifying them as static. The process of the DFM is as follows: Initially, we calculate the average depth $\mu_{d}$ of all feature points within the bounding box, provided there are a sufficient number of feature points. Subsequently, we calculate the standard deviation $\delta_{d}$ of these points. Any feature points falling outside the range $[\mu_{d}-\eta  \delta_{d}, \mu_{d} + \eta  \delta_{d}]$ are eliminated from the potential dynamic feature point set and treated as static in coarse estimation stage. Here, $\eta$ represents a constant value that describes the degree of deviation from the average depth value.

\subsection{Coarse Estimation Module}
Stereo images are used for ORB feature extraction initially. The left image is utilized for object detection and tracking using YOLOv7 \cite{wang2023yolov7} and Deep Sort \cite{du2023strongsort}. In comparison to CFP-SLAM, which uses an extended Kalman filter (EKF) and Hungarian algorithm to compensate for missed object detection, we employ Deep Sort for a more effective implementation. Leveraging class information from the object detection model, we utilize the DFM to treat background features as static points and combine them with all static feature points outside of the bounding boxes to estimate the camera pose by minimizing the re-projection error as in equation (\ref{eq:minerror}).

\subsection{Multilayer Perceptron Module}
This section details the construction and deployment of a MLP model to effectively differentiate between dynamic and static feature points.

Before deploying the MLP model for classifying feature points, a sufficient training dataset of dynamic and static points is collected. Details of the dataset collection process are provided in Section \ref{subsec:data_col}. Once an adequate dataset is available, the structure of the MLP model needs to be designed. The MLP model consists of three parts: the input layer, the hidden layer, and the output layer. The input layer's number of nodes corresponds to the dimension of the input feature, which is three in this case:  the image intensity error $e_I$, the epipolar
error $e_D$, and the re-projection error $e_{Re}$. Previous work like\cite{CFP-SLAM} and \cite{PointSLOT} has often overlooked some of these features, leading to incomplete information. 

\begin{figure}
      \centering
      \includegraphics[width=0.5\textwidth, height=0.15\textheight]{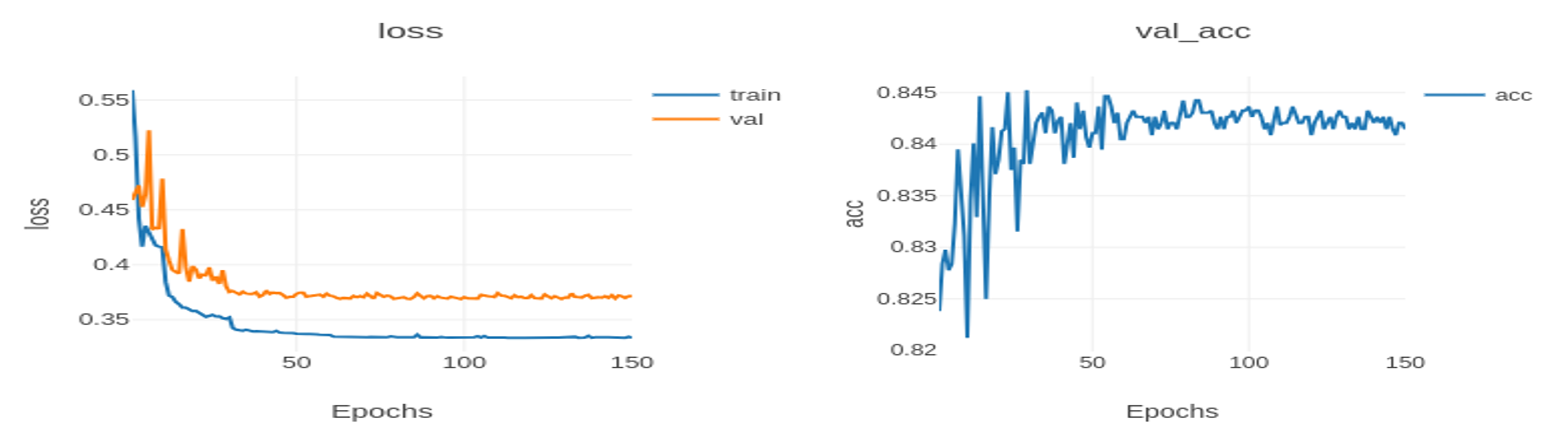}
      \caption{Training process of optimal MLP model. It is showing a consistent decrease in both training loss and validation losses indicated successful learning.Also, the Accuracy metric, which calculates the ratio between the number of correct classification cases and the number of total cases, is shown to be increasing. }
      \label{figurelabel3}
   \end{figure}

Training is crucial for classifying dynamic and static feature points. The process begins with zero-mean normalization of the datasets, followed by dividing the dataset into training, testing, and validation sets. Before training, hyperparameters and optimizer settings are configured as detailed in Section \ref{subsec:exp_set}. After lots of experiments, the optimal MLP model training process is presented in Figure \ref{figurelabel3}, showing a consistent decrease in both training loss and validation losses indicated successful learning. The Accuracy metric, which calculates the ratio between the number of correct classification cases and the number of total cases, is shown to be increasing.


The MLP prediction module begins by generating input feature $v_i$ for each feature point $x_i$ that is located in the bounding box and is not filtered by DFM, then normalizing the input feature $v_i$ through zero mean normalization to eliminate dimension effects, using the same parameters as the dataset normalization. Finally, all input features are put into the model in batches of a predetermined size and are classified as static or dynamic.

\subsection{Fine Estimation}  \label{subsec:fine_est}
Upon utilizing the MLP model to classify feature points, we are able to differentiate between static and dynamic feature points located within the bounding box of all objects. Subsequently, feature points outside the bounding box, as well as static points filtered by the Depth Filter and classified as static by the MLP, are combined to form a comprehensive set of static feature points. During the fine estimation phase, the camera pose initially generated in the coarse estimation stage is refined by minimizing the re-projection error of the static feature points.

\subsection{Data Collection And Data Split} \label{subsec:data_col}
At present, all methods for classifying dynamic and static feature points are evaluated indirectly through SLAM metrics like absolute translation error (ATE), relative translation error (RTE) and relative rotation error (RRE), primarily due to the absence of a dedicated classification dataset. This reliance on indirect assessment may result in inaccuracies and instability, as it fails to account for the inherent randomness present in SLAM systems. By establishing a comprehensive classification dataset, it becomes possible to evaluate dynamic and static feature point classification methods directly, thereby enhancing the reliability of experimental results. For bridging this gap, we build a publicly available datasets which contains over 50,000 dynamic and static feature points from sequences 00 and 01 of KITTI Odometry datasets. All those feature points manually labeled, and the ratio of dynamic and static feature points in collected dataset is 1:1. Each feature points is saved as following format, $u_1$, $v_1$, $z_1$, $id_1$, $u_2$, $v_2$, $id_2$, $class$, $e_I$, $e_{Re}$, $e_D$. The ($u_1$, $v_1$), ($u_2$, $v_2$) stands for coordinates of feature point $x_1$ and $x_2$ in image frame respectively. And $z_1$ means the depth of $x_1$, which is computed through the same way as in ORB-SLAM2. The $id_1$ and $id_2$ represent whether the image frames belong to $x_1$ or $x_2$. The $class$ is binary, either $dynamic$ or $static$. The most important properties are $e_I$, $e_{Re}$ and $e_D$, and they represent image intensity error, re-projection error and epipolar error respectively. We use ground truth pose of each frame to generated them. Details of how they are computed is represented in Eq. (\ref{eq:error1}), (\ref{eq:error2}) and (\ref{eq:error3}). 

A proper dataset split is crucial for training supervised machine learning models. We shuffle the dataset of over 50,000 feature points at first, then allocate 70\% as training set, 20\% as testing set, and 10\% as validation set. Besides, We maintain a 1:1 ratio of static to dynamic feature points in all sets to address data imbalance.

\section{Experimental Results}
\subsection{Experimental Settings}  \label{subsec:exp_set}
A C++-implemented MLP-based real-time stereo SLAM system is tested on a laptop with an Intel Core i9-13900k@5.8 GHz processor. The YOLOv7 and DeepSort are used to detect object class, locate 2D bounding boxes, and track objects. Both of them are executed on an Nvidia RTX3080 GPU, and the rest of the SLAM system runs on the CPU.

Parameter $\eta$ in the DFM is set to $1.2$. Parameters related to the training part of the MLP model are as follows: the \textit{epoch} is 150, the \textit{batch size} is 8, the \textit{learning rate} is 0.1, the \textit{learning rate scheduler} is \textit{MultiStepLR} \cite{paszke2019pytorch} with $\gamma$ set to 0.1. The optimizer we choose is \textit{Adam} \cite{kingma2014adam}. We converted the TensorRT \cite{tensorRT} version of the MLP model with a fixed input shape of (200, 3).

\begin{table}[ht]
\caption{Classification Comparison}
\label{table1}
\begin{center}
\begin{tabular}{ccc}
\toprule
Method & Accuracy & F1-Score \\
\toprule
MLP & \textbf{87.71} & \textbf{87.64}\\
SVM & 84.19 & 84.58\\ 
method similar to PointSLOT & 81.36 & 81.43\\
method similar to CFP-SLAM & 73.61 & 71.10 \\
$e_{D}$ only & 60.10 & 61.32\\
\hline
\end{tabular}
\end{center}
\end{table}

\subsection{Classification Evaluation}
We have compiled a dataset to classify dynamic and static feature points, consisting of over 50,000 feature points. This dataset enables the direct evaluation of various dynamic and static feature classification methods, eliminating the need for indirect comparisons based on metrics from the SLAM field. Direct evaluation is advantageous as it is less influenced by other factors within the SLAM system, resulting in more reliable and accurate experimental outcomes.

Our experiment involves the assessment of various methods for feature discrimination, including our MLP based dynamic and static feature discriminator, Support Vector Machine (SVM), a method akin to PointSLOT, a method similar to CFP-SLAM, and a method focusing solely on epipolar error. SVM is a well-established classification technique in machine learning, widely used in text and hypertext categorization, image classification, and satellite data classification. In this study, SVM from the Scikit-learn library \cite{pedregosa2011scikit} is directly employed, trained, and evaluated on the dataset. Additionally, two classification methods for dynamic and static feature points, similar to PointSLOT and CFP-SLAM, are implemented in Python due to the unavailability of their source code. The MLP based method and SVM are trained on the same training set, and all five methods are tested on the same testing set. The training set, testing set and validation set are obtained through the data split method in Section \ref{subsec:data_col}.  Experimental results presented in Table \ref{table1} are the best performance in the testing set of collected dataset by fine-tuning the hyper-parameters of the methods. The PointSLOT solely focuses on re-projection error, while the CFP-SLAM considered epipolar error in addition. Furthermore, a discrimination method concentrating solely on epipolar error is also implemented. The evaluation of these methods is based on the metrics of Accuracy and F1-score. Accuracy reflects the model's predictive correctness, with true positive (TP), false positive (FP), true negative (TN), and false negative (FN) used to denote classification outcomes.. And the Accuracy is described as following:
\begin{equation}
    \text{Accuracy}=\frac{TP + TN}{TP + FP + TN + FN}
\end{equation}
The F1-Score is a measure of the predictive performance of the model. It is the harmonic mean of the Precision and Recall. The Precision and Recall indicated correct identified proportion of positive identifications and correct proportion of actual positives respectively. Thus, the F1-Score symmetrically represents both precision and recall in one metric. It can be described as follows

\begin{equation}
\text{F1-Score}=\frac{2 \times \text{Precision} \times \text{Recall}}{\text{Precision} + \text{Recall}}
\end{equation}
\begin{equation}
\text{Precision}=\frac{TP}{TP+FP}
\end{equation}
\begin{equation}
\text{Recall}=\frac{TP}{TP+FN}
\end{equation}

The experimental results are presented in Table \ref{table1}. According to it, we can see that our MLP-based method achieves superior performance in terms of both Accuracy and F1-Score compared to other methods. Additionally, the support-vector machine (SVM) outperforms other methods except for our MLP-based method, indicating that machine learning-based methods are more suitable for this dynamic and static feature classification problem compared to traditional methods that require manual parameter tuning. The significant difference between the method that only considers epipolar error and the method similar to PointSLOT which only considers re-projection error, suggests that re-projection error is a more critical indicator than epipolar error for this problem.

\begin{table*}[ht]
\caption{The ATE Comparison on KITTI Datasets}
\label{table_2}
\begin{center}
\begin{tabular}{cccccc}
\toprule
\textbf{Sequences} & \textbf{ORB-SLAM2} & \textbf{DynaSLAM} & \textbf{DynaSLAM II} & \textbf{VDO-SLAM} & \textbf{Ours}\\
\midrule
00 & 0.91 & 1.40 & 1.29 & -- & \textbf{0.79}\\
01 & 7.35 & 9.40 & \textbf{2.31} & 189.13 & 4.53\\
02 & 3.36 & 6.70 & \textbf{0.91} & -- & 3.48 \\
03 & 0.39 & 0.60 & 0.69 & 5.36 & \textbf{0.34}\\
04 & \textbf{0.15} & 0.20 & 1.42& 3.46 & 0.24\\
05 & 0.54 & 0.80 & 1.34 & -- & \textbf{0.33}\\
06 & 0.44 & 0.80 & \textbf{0.19} & 3.63 & 0.88\\
07 & 0.42& 0.50 & 3.30 & 2.73 & \textbf{0.41}\\
08 & 3.03 & 3.50 & \textbf{1.68} & -- & 3.18\\
09 & 3.97 & 1.60 &5.02 & 11.73 & \textbf{0.90}\\
10 & 1.34 & 1.30 & 1.30 & 3.72 & \textbf{1.27}\\
\midrule
\textbf{average} & 1.99 & 2.43 & 1.77 & -- & \textbf{1.48} \\
\bottomrule
\end{tabular}
\end{center}
\end{table*}

\begin{table}[ht]
\caption{The ATE Comparison on KITTI360 Datasets}
\label{table_extra}
\begin{center}
\begin{tabular}{cccccc}
\toprule
\textbf{Sequences} & \textbf{ORB-SLAM2} & \textbf{VDO-SLAM} & \textbf{Ours}\\
\midrule
00 & 1.46 & 6.92  & \textbf{1.36}\\
01 & 2.83 & 10.73  & \textbf{2.04}\\
02 & 6.82 & --  & \textbf{6.17 }\\
03 & 4.58 & --  & \textbf{4.11}\\
\midrule
\textbf{average} & 3.92 & 8.83 & \textbf{3.42} \\
\bottomrule
\end{tabular}
\end{center}
\end{table}

\subsection{Visual Odometry}
KITTI Odometry dataset is a publicly available collection of stereo visual odometry data, with 11 training and 10 testing sequences featuring urban and road scenes from a car's perspective, including moving vehicles and GPS data. KITTI360 enhances this dataset with better localization and richer semantic information.

We conduct a comprehensive comparison between our MLP-based SLAM system and other SLAM systems in this section. A presentation video of our experiment can be found in \href{https://www.youtube.com/watch?v=sOKNc0-ZH2Q}{https://www.youtube.com/watch?v=sOKNc0-ZH2Q}. In Table \ref{table_2}, we provide an absolute translation error (ATE) comparison with ORB-SLAM2, DynaSLAM, DynaSLAM II, and a modified version of VDO-SLAM on the KITTI Odometry dataset. It is important to note that we deleted portions of the image frames in sequences 00 and 01 of KITTI Odometry dataset which were used for collecting the training dataset for the MLP model. Moreover, comparison with ORB-SLAM2, DynaSLAM, modified VDO-SLAM on the KITTI360 dataset is presented in Table \ref{table_extra}. The MLP model for distinguishing dynamic and static feature points was trained only on the KITTI Odometry dataset and did not use any possible information from the KITTI360 dataset. ORB-SLAM2 does not specifically address dynamic situations but serves as the foundation of our SLAM system. DynaSLAM also builds upon ORB-SLAM2 with the ability to detect dynamic objects, but it discards features located within the bounding box of potential dynamic objects directly, leading to information loss. DynaSLAM II is the current state-of-the-art method, aiming to enhance camera pose estimation precision by optimizing the Bundle Adjustment (BA) factor graph, which includes object poses and constant velocity constraints. The modified version of VDO-SLAM \cite{vdo-slam-modified} is a re-implemented version; the original version can only run on datasets provided by the authors. The modified VDO-SLAM requires acquiring depth images through ZoeDepth \cite{bhat2023zoedepth}, optical flow through VideoFlow offline, and applying YOLOv8 \cite{Jocher_Ultralytics_YOLO_2023} to detect dynamic objects. We tested all SLAM systems on our experimental devices except DynaSLAM II due to the unavailability of its source code. Therefore, we directly used the experimental results from its original paper.

According to experimental results in Table \ref{table_2} and Table \ref{table_extra}, it can be seen that our SLAM system outperforms other SLAM system in most of sequences of KITTI Odometry and KITTI360 dataset. More specifically, our SLAM system outperforms VDO-SLAM-modified version in all sequences. Besides, compared to our base ORB-SLAM2, we achieve better performance in the sequences 00, 01, 03, 05, 07, 08, 09 and 10 of KITTI Odometry dataset and in all sequences of KITTI360 dataset. The sequence 01 was collected from a car on the highway, featuring dynamic vehicles. Our SLAM system surpasses ORB-SLAM2 in this sequence, proving its ability to effectively manage dynamic objects in the environment. Moreover, compared to DynaSLAM, our SLAM system achieves higher precision on sequences except for sequences 04 and 06 of KITTI Odometry dataset, and the gap on the sequence 04 and 06 are within 0.1. Compared to DynaSLAM II, our SLAM system only performs worse on the sequences 01, 02, 06 and 08 out of 11 sequences. Our SLAM system recorded the lowest average absolute translation error (ATE) on the KITTI Odometry and KITTI360 datasets, demonstrating the highest average precision across all sequences. It is worth mentioning that the sequence 00 is rich of static cars that are parked on the roadside, our SLAM system achieves the best performance in this sequence compared to other methods, which indicates that our MLP model is capable to identifying real dynamic objects. Although experiments are conducted solely using a stereo sensor, our proposed method can be extended to accommodate monocular or RGB-D cameras.

\begin{table}[ht]
\caption{Latency Of Each Module}
\label{table_3}
\begin{center}
\begin{tabular}{c|cc}
\toprule
Module & ORB-SLAM2 (stereo) & Ours \\
\midrule
Feature Process & 37.31 ms & 37.61 ms \\
Depth Filter & -- ms & 2.32 ms \\
Pose Estimation & 1.78 ms & 2.58 ms\\
YOLOv7 Detection &  --& 7.72 ms\\
Deep Sort Tracking &  --& 4.66 ms\\
MLP Inference &  --& \textbf{0.18} ms\\
Other Modules & 6.22 ms & 7.74 ms \\
Tracking & 45.31 ms & 62.81 ms\\
\bottomrule
\end{tabular}
\end{center}
\end{table}

\subsection{Time Analysis}
Low latency is one of the desired property of the SLAM system. We lists the average computational time of each module in the tracking thread of our MLP based stereo SLAM system in Table \ref{table_3}. According to experimental results from Table \ref{table_2} and Table \ref{table_3}, our MLP-based SLAM system achieves higher precision with only 17.50 ms/frame extra computation cost compared to ORB-SLAM2. Moreover, we made a comprehensive latency comparison with other SLAM systems. The mean values of the highest latency and the lowest latency from the original papers of DynaSLAM II and PointSLOT are presented in Table \ref{table_4}. For all other methods in Table \ref{table_4}, average latency value in KITTI Odometry sequences from 00 to 10 are presented. 
It is worth mentioning that even though the latency of DynaSLAM II does not include the time consumption of 2d and 3d object detection models, the execution time of our methods is still 17.34 ms faster. Combining Table \ref{table_2}, Table \ref{table_extra} and Table \ref{table_4}, we can conclude that our MLP-based SLAM system achieves the lowest average absolute translation error (ATE) and the fastest speed compared to other dynamic SLAM systems.

\begin{table}[h]
\caption{Latency Anaylysis}
\label{table_4}
\begin{center}
\begin{tabular}{ccc}
\toprule
Method & Sensor & Latency\\
\midrule
ORB-SLAM2 & Stereo & 45.31 ms (base)\\
DynaSLAM & Stereo & 519.37 ms\\
DynaSLAM II & Stereo & 80.15 ms\\
VDO-SLAM & Stereo + Dpeth & 86.97 ms\\
PointSLOT &  Stereo & 73.6 ms \\
Ours & Stereo & \textbf{62.81 ms}\\
\bottomrule
\end{tabular}
\end{center}
\end{table}

\subsection{Ablation Study}
We evaluated our MLP module's effectiveness for dynamic feature point removal by comparing three models.Our SLAM system features just the object tracking module. Our SLAM system uses only the MLP module.And SLAM system features both modules and additional components stayed the same. The experimental results are presented in Table \ref{table_5}. It indicated that the SLAM system with MLP-only module outperforms the object tracking module, and their combination achieves the best performance.

\begin{table}[ht]
\caption{The ATE Comparison on KITTI Datasets of Abalation Study}
\label{table_5}
\begin{center}
\begin{tabular}{cccccc}
\toprule
\textbf{Sequences} & \textbf{object tracking only} & \textbf{MLP only} & \textbf{both} \\
\midrule
00 & 2.53 & 1.37 & \textbf{0.79} \\
01 & 10.71 & 6.79 & \textbf{4.53}\\
02 & 5.73 & 4.01 & \textbf{3.48} \\
03 & 0.42 & \textbf{0.34} & \textbf{0.34}\\
04 & 0.18 & \textbf{0.17 }& 0.24\\
05 & 0.38 & 0.35 & \textbf{0.33}\\
06 & 1.03 & \textbf{0.80} & 0.88\\
07 & 0.49 & 0.43 & \textbf{0.41}\\
08 & 3.36 &  \textbf{2.82} & 3.18\\
09 & 3.43 & 3.31 & \textbf{0.90}\\
10 & \textbf{1.16} & 1.23 & 1.27\\
\midrule
\textbf{average} & 2.67 & 1.97 & \textbf{1.48} \\
\bottomrule
\end{tabular}
\end{center}
\end{table}

\section{CONCLUSIONS}

In this letter, we propose a MLP-based real-time stereo SLAM system to address SLAM system degradation in dynamic environments. Currently, there is no publicly available dataset for directly evaluating the effectiveness of dynamic and static feature classification methods. Previous assessments have relied on indirect metrics, which are prone to inaccuracy due to the randomness in the Random Sample Consensus (RANSAC) process in camera pose estimation within the SLAM systems. To bridge this gap, we have created a publicly available dataset containing over 50,000 feature points. Our proposed MLP-based dynamic and static feature discriminator is the most effective method compared to other methods on our collected datasets. Additionally, our MLP-based real-time stereo SLAM system achieves the highest average precision on the KITTI Odometry dataset and the fastest speed compared to other dynamic SLAM systems. 

\addtolength{\textheight}{-12cm}   





\section*{ACKNOWLEDGMENT}
We would like to acknowledge the subaward from the Virginia Tech under the U.S. Army Grant No. W911QX-23-2-0001.

\bibliographystyle{IEEEtran}
\bibliography{references}
\end{document}